\documentclass[11pt]{article}
\usepackage{acl2016}
\usepackage{times}
\usepackage{url}
\usepackage{latexsym}

\usepackage{multirow}

\usepackage{stmaryrd} 

\usepackage{amsmath,amssymb}
\usepackage{graphicx}

\aclfinalcopy 
\def\aclpaperid{511} 

\newcommand\attent{\stackrel{a}{\shortrightarrow}}

\title{Using Sentence-Level LSTM Language Models for Script Inference}

\author{Karl Pichotta \\
  Department of Computer Science \\
  The University of Texas at Austin \\
  {\tt pichotta@cs.utexas.edu} \\\And
  Raymond J. Mooney \\
  Department of Computer Science \\
  The University of Texas at Austin \\
  {\tt mooney@cs.utexas.edu} \\}

\date{}

\begin{document}
\maketitle
\begin{abstract}
  There is a small but growing body of research on statistical scripts, models
  of event sequences that allow probabilistic inference of
  implicit events from documents. These systems operate on structured
  verb-argument events produced by an NLP pipeline. We compare these systems
  with recent Recurrent Neural Net models that directly operate on raw tokens to
  predict sentences, finding the latter to be roughly comparable to the former
  in terms of predicting missing events in documents.
\end{abstract}

\section{Introduction}

Statistical scripts are probabilistic models of event sequences
\cite{chambers:acl08}. A learned script model is capable of
processing a document and inferring events that are probable but not explicitly stated.
These models operate on automatically extracted structured events (for
example, verbs with entity arguments), which are derived from standard NLP tools
such as dependency parsers and coreference resolution engines.

Recent work has demonstrated that standard sequence models applied to such extracted
event sequences, e.g.\ discriminative language models
\cite{rudinger:emnlp15} and Long Short Term Memory (LSTM) recurrent neural nets
\cite{pichotta:aaai16}, are able to infer held-out events more accurately than previous approaches.
These results call into question the extent to which
statistical event inference systems require linguistic preprocessing and 
syntactic structure. In an
attempt to shed light on this issue, we compare existing script models to LSTMs trained as {\it sentence-level language models} which try to predict the sequence of words in the next sentence from a learned representation of the previous sentences using no linguistic preprocessing.

Some prior statistical script learning systems are focused on knowledge 
induction. These systems are primarily designed to induce collections of 
co-occurring event types involving the same entities, and their ability to infer
held-out events is not their primary intended purpose 
\nocite{chambers:acl08,ferraro:aaai16} (Chambers and Jurafsky, 2008; Ferraro and 
Van Durme, 2016, \textit{inter alia}). In the present work, we instead 
investigate the behavior of systems trained to directly optimize performance on 
the task of predicting subsequent events; in other words, we are investigating 
statistical models of events in discourse.

Much prior research on statistical script learning has also evaluated on
inferring missing events from documents. However, the exact form that this task 
takes depends on the adopted definition of what constitutes an event: in 
previous work, events are defined in different ways, with differing degrees of 
structure. We consider simply using raw text, which requires no explicit 
syntactic annotation, as our mediating representation, and evaluate how raw text 
models compare to models of more structured events.

\newcite{kiros:nips15} introduced \textit{skip-thought vector} models, in which
an RNN is trained to encode a sentence within a document
into a low-dimensional vector that supports predicting the neighboring
sentences in the document. Though the objective function used to train networks
maximizes performance on the task of predicting sentences from their neighbors,
\newcite{kiros:nips15} do not evaluate directly on the ability of networks to
predict text; they instead demonstrate that the intermediate low-dimensional
vector embeddings are useful for other tasks. We directly evaluate the text
predictions produced by such sentence-level RNN encoder-decoder models, and
measure their utility for the task of predicting subsequent events.

We find that, on the task of predicting the text of held-out sentences,
the systems we train to operate on the level of raw text generally outperform
the systems we train to predict text mediated by automatically extracted event
structures. On the other hand, if we run an NLP pipeline on the automatically
generated text and extract structured events from these predictions, we achieve
prediction performance roughly comparable to that of systems trained to predict
events directly. The difference between word-level and
event-level models on the task of event prediction is marginal, indicating that the task of predicting the next
event, particularly in an encoder-decoder setup, may not necessarily need to be 
mediated by explicit event structures.
To our knowledge, this is the first effort to evaluate sentence-level RNN
language models directly on the task of predicting document text. Our results show
that such models are useful for predicting missing information in text; and the
fact that they require no linguistic preprocessing makes them more applicable to languages where
quality parsing and co-reference tools are not available.

\section{Background}

\subsection{Statistical Script Learning}

\textit{Scripts}, structured models of stereotypical sequences of
events, date back to AI research from the 1970s, in particular the seminal
work of \newcite{schank:book77}. In this conception, scripts are modeled as temporally ordered sequences of symbolic structured events. These models are nonprobabilistic and brittle, and pose
serious problems for automated learning.

In recent years, there has been a growing body of research into statistical
script learning systems, which enable statistical inference of implicit events
from text. \nocite{chambers:acl08,chambers:acl09} Chambers and Jurafsky (2008;
2009) describe a number of simple event co-occurrence based systems which infer
(verb, dependency) pairs related to a particular discourse entity. For example,
given the text:
\begin{quote}
\textit{Andrew Wiles won the 2016 Abel prize for proving Fermat's
last theorem},
\end{quote}
such a system will ideally be able to infer novel facts like
(\textit{accept}, subject) or (\textit{publish}, subject) for the entity
\textit{Andrew Wiles}, and facts like (\textit{accept}, object) for the entity
\textit{Abel prize}. A number of other systems inferring the same types of
pair events have been shown to provide superior performance in modeling events
in documents \cite{jans:eacl12,rudinger:emnlp15}.

\newcite{pichotta:eacl14} give a co-occurrence based script system
that models and infers more complex multi-argument events from text. For
example, in the above example, their model would ideally be able to infer a
single event like \textit{accept}(\textit{Wiles}, \textit{prize}), as opposed
to the two simpler pairs from which it is composed. They provide evidence that
modeling and inferring more complex multi-argument events also yields superior
performance on the task of inferring simpler (verb, dependency) pair events.
These events are constructed using only coreference information; that is, the
learned event co-occurrence models do not directly incorporate noun information.

More recently, \newcite{pichotta:aaai16} presented an LSTM-based script
inference model which models and infers multi-argument events, improving on
previous systems on the task of inferring verbs with arguments. This system can
incorporate both noun and coreference information about event arguments. We will
use this multi-argument event formulation (formalized below) and compare LSTM models using this event formulation to LSTM models using raw text.

\subsection{Recurrent Neural Networks}
\label{sec:rnn}

Recurrent Neural Networks (RNNs) are neural nets whose computation graphs
have cycles. In particular, RNN sequence models are RNNs which map a sequence
of inputs $x_1, \ldots, x_T$ to a sequence of outputs $y_1, \ldots, y_T$ via
a learned latent vector whose value at timestep $t$ is a function of
its value at the previous timestep $t-1$.

The most basic RNN sequence models, so-called ``vanilla RNNs''
\cite{elman:cogscij90}, are
described by the following equations:
\begin{align*}
z_t &= f(W_{i,z} x_t + W_{z,z} z_{t - 1}) \\
o_t &= g(W_{z,o} z_t)
\end{align*}
where $x_t$ is the vector describing the input at
time $t$; $z_t$ is the vector giving the hidden
state at time $t$; $o_t$ is the vector giving the predicted
output at time $t$;
$f$ and $g$ are element-wise nonlinear functions (typically sigmoids,
hyperbolic tangent, or rectified linear units);
and $W_{i,z}$, $W_{z,z}$, and $W_{z,o}$ are learned
matrices describing linear transformations. The recurrency
in the computation graph arises from the fact that $z_t$ is a function of
$z_{t-1}$.

\begin{figure}[t]
\centering
\includegraphics[width=0.6\textwidth]{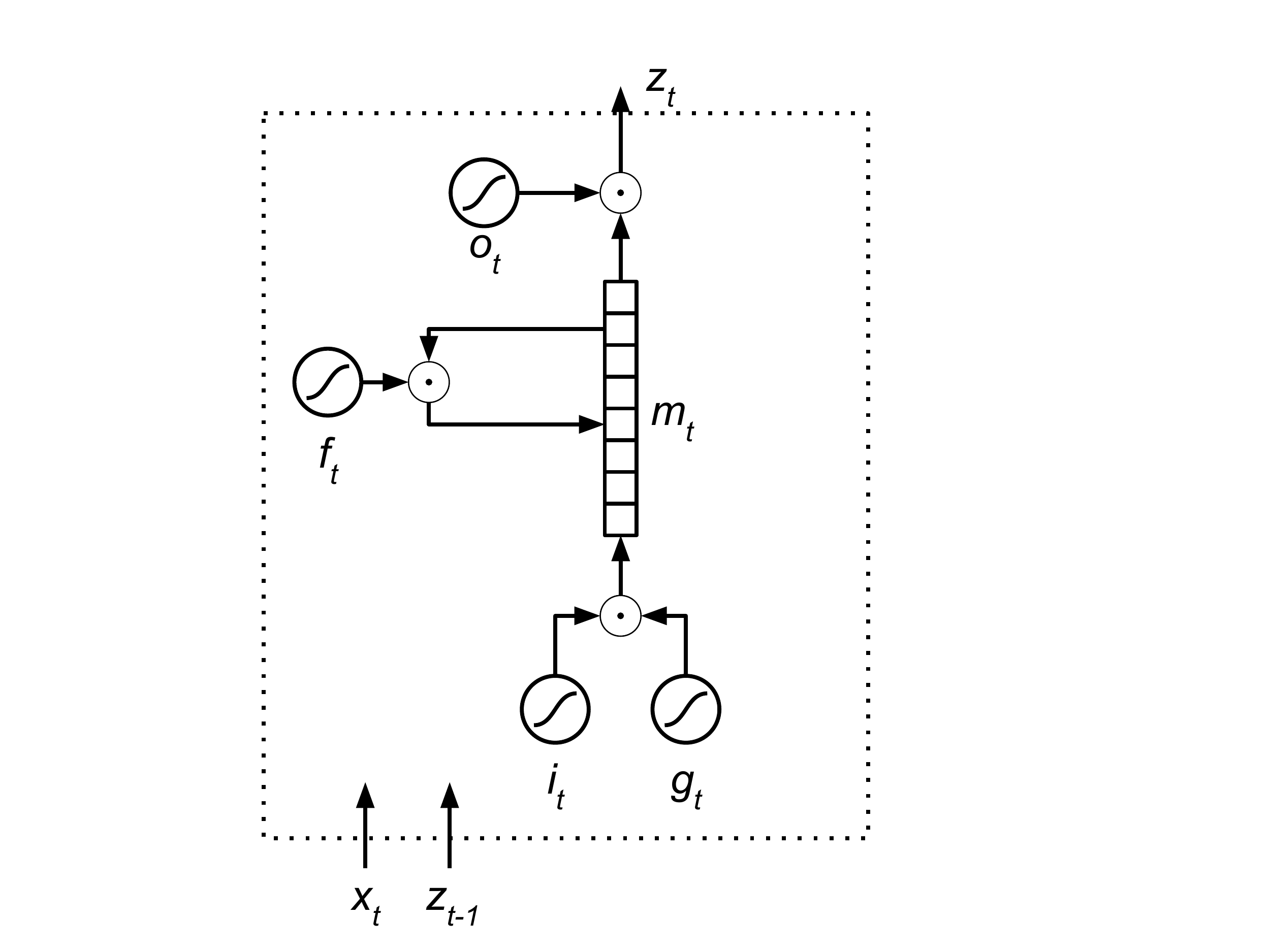}
\caption{Long Short-Term Memory unit at timestep $t$. The
four nonlinearity nodes ($i_t$, $g_t$, $f_t$, and $o_t$) all have, as inputs,
$x_t$ and $z_{t-1}$. Small circles with dots are elementwise vector
multiplications.
}
\label{fig:lstm}
\end{figure}

The more complex
Long Short-Term Memory (LSTM) RNNs \cite{hochreiter:nc97} have
 been shown to perform well on a wide variety of NLP tasks
\nocite{sutskever:nips14,hermann:nips15,vinyals:nips15}
(Sutskever et al., 2014; Hermann et al., 2015; Vinyals et al., 2015,
\textit{inter alia}).
The LSTM we use is described by:
\begingroup
\allowdisplaybreaks
\begin{align*}
i_t &= \sigma\left( W_{x,i} x_t + W_{z,i} z_{t - 1} + b_i    \right) \\
f_t &= \sigma\left( W_{x,f} x_t + W_{z,f} z_{t - 1} + b_f    \right) \\
o_t &= \sigma\left(  W_{x,o} x_t + W_{h,i} z_{t - 1} + b_o   \right) \\
g_t &= \tanh\left(   W_{x,m} x_t + W_{z,m} z_{t - 1} + b_g   \right) \\
m_t &= f_t \circ m_{t-1} + i_t \circ g_t \\
z_t &= o_t \circ \tanh m_t.
\end{align*}
\endgroup
The model is depicted graphically in Figure \ref{fig:lstm}. The memory
vector $m_t$ is a function of both its previous value $m_{t-1}$ and the input
$x_t$; the vector $z_t$ is output both to any layers above the unit (which
are trained to predict the output values $y_t$), and is additionally given as
input to the LSTM unit at the next timestep $t+1$. The $W_{*,*}$
matrices and $b_{*}$ vectors are learned model parameters, and
$u\circ v$ signifies element-wise multiplication.

\subsection{Sentence-Level RNN Language Models}
\label{sec:sentrnn}

RNN sequence models have recently been shown to be extremely effective
for word-level and character-level language models
\cite{mikolov:interspeech11,jozefowicz:arxiv16}. At each timestep, these models
take a word or character as input, update a hidden state vector, and predict
the next timestep's word or character. There is also a growing body of work on
training RNN encoder-decoder models for NLP problems. These systems first encode
the entire input into the network's hidden state vector and then, in a second
step, decode the entire output from this vector
\cite{sutskever:nips14,vinyals:nips15,serban:aaai16}.

Sentence-level RNN language models, for example the skip-thought vector system
of \newcite{kiros:nips15}, conceptually bridge these two approaches.  Whereas 
standard language models are trained to predict the next token in the sequence 
of tokens, these systems are explicitly trained to predict the next 
{\it sentence} in the sequence of sentences.
\newcite{kiros:nips15} train an encoder-decoder model to encode a sentence into
a fixed-length vector and subsequently decode both the following and preceding
sentence, using Gated Recurrent Units \cite{chung:nips14}. In the present work,
we train an LSTM model to predict a sentence's successor, which is essentially
the forward component of the skip-thought system. \newcite{kiros:nips15} use
the skip-thought system as a means of projecting sentences into low-dimensional
vector embeddings, demonstrating the utility of these embeddings on a number of
other tasks; in contrast, we will use our trained sentence-level RNN language
model directly on the task its objective function optimizes: predicting a
sentence's successor.

\section{Methodology}

\subsection{Narrative Cloze Evaluation}
\label{sec:cloze}

The evaluation of inference-focused statistical script systems is not
straightforward. \newcite{chambers:acl08} introduced the \textit{Narrative
Cloze} evaluation, in which a single event is held out from a document and
systems are judged by the ability to infer this held-out event given the
remaining events. This evaluation has been used by a number of published script
systems \cite{chambers:acl09,jans:eacl12,pichotta:eacl14,rudinger:emnlp15}. This
automated evaluation measures systems' ability to model and predict events as they
co-occur in text.

The exact definition of the Narrative Cloze evaluation depends on the
formulation of events used in a script system. For example,
\newcite{chambers:acl08}, \newcite{jans:eacl12}, and \newcite{rudinger:emnlp15}
evaluate inference of held-out (verb, dependency) pairs from
documents; \newcite{pichotta:eacl14} evaluate inference of verbs with
coreference information about multiple arguments; and \newcite{pichotta:aaai16} evaluate
inference of verbs with noun information about multiple arguments. In order to gather
human judgments of inference quality, the latter also learn an
encoder-decoder LSTM network for transforming verbs and noun arguments into
English text to present to annotators for evaluation.

We evaluate instead on the task of directly inferring sequences of words. That
is, instead of defining the Narrative Cloze to be the evaluation of predictions
of held-out events, we define the task to be the evaluation of predictions of
held-out text; in this setup, predictions need not be mediated by noisy,
automatically-extracted events. To evaluate inferred text against gold standard
text, we argue that the \textsc{bleu} metric \cite{papineni:acl02}, commonly
used to evaluate Statistical Machine Translation systems, is a natural
evaluation metric. It is an n-gram-level analog to the event-level Narrative
Cloze evaluation: whereas the Narrative Cloze evaluates a system on its ability
to reconstruct events as they occur in documents, \textsc{bleu} evaluates a
system on how well it reconstructs the n-grams. 

This evaluation takes some
inspiration from the evaluation of neural encoder-decoder translation models
\cite{sutskever:nips14,bahdanau:iclr15}, which use similar architectures for 
the task of Machine Translation. That is, the task we present can be thought of 
as ``translating'' a sentence into its successor.
While we do not claim that \textsc{bleu} is necessarily the optimal way of 
evaluating text-level inferences, but we do claim that it is a natural 
ngram-level analog to the Narrative Cloze task on events.

If a model infers text, we may also evaluate it on the task of inferring
events by automatically extracting structured events from its output text (in
the same way as events are extracted from natural text). This allows us to
compare directly to previous event-based models on the task they are optimized
for, namely, predicting structured events.

\subsection{Models}

Statistical script systems take a sequence of events from a document and infer
additional events that are statistically probable. Exactly what constitutes an
\textit{event} varies: it may be a (verb, dependency) pair inferred as relating
to a particular discourse entity \cite{chambers:acl08,rudinger:emnlp15}, a
simplex verb \cite{chambers:acl09,orr:aaai14}, or a verb with multiple
arguments \cite{pichotta:eacl14}. In the present work, we adopt a representation
of events as verbs with multiple arguments
\cite{balasubramanian:emnlp13,pichotta:eacl14,modi:conll14}. Formally, we define
an event to be a variadic tuple $(v, s, o, p^*)$, where $v$ is a verb, $s$ is
a noun standing in subject relation to $v$, $o$ is a noun standing as a direct
object to $v$, and $p^*$ denotes an arbitrary number of (\textit{pobj},
\textit{prep}) pairs, with \textit{prep} a preposition and \textit{pobj} a noun
related to the verb $v$ via the preposition \textit{prep}.\footnote{This is
essentially the event representation of \newcite{pichotta:aaai16}, but whereas
they limited events to having a single prepositional phrase, we allow an
arbitrary number, and we do not lemmatize words.}
Any argument except $v$ may be \textit{null}, indicating no noun fills that
slot. For example, the text
\begin{quote}
  \textit{Napoleon sent the letter to Josephine}
\end{quote}
would be represented by the event (\textit{sent, Napoleon, letter, (Josephine,
to)}). We represent arguments by their grammatical head words.

We evaluate on a number of different neural models which differ in their
input and output. All models are LSTM-based encoder-decoder models. These models
encode a sentence (either its events or text) into a learned hidden vector state
and then, subsequently, decode that vector into its successor sentence (either
its events or its text).

\begin{figure*}[t]
\centering
\includegraphics[width=0.8\textwidth]{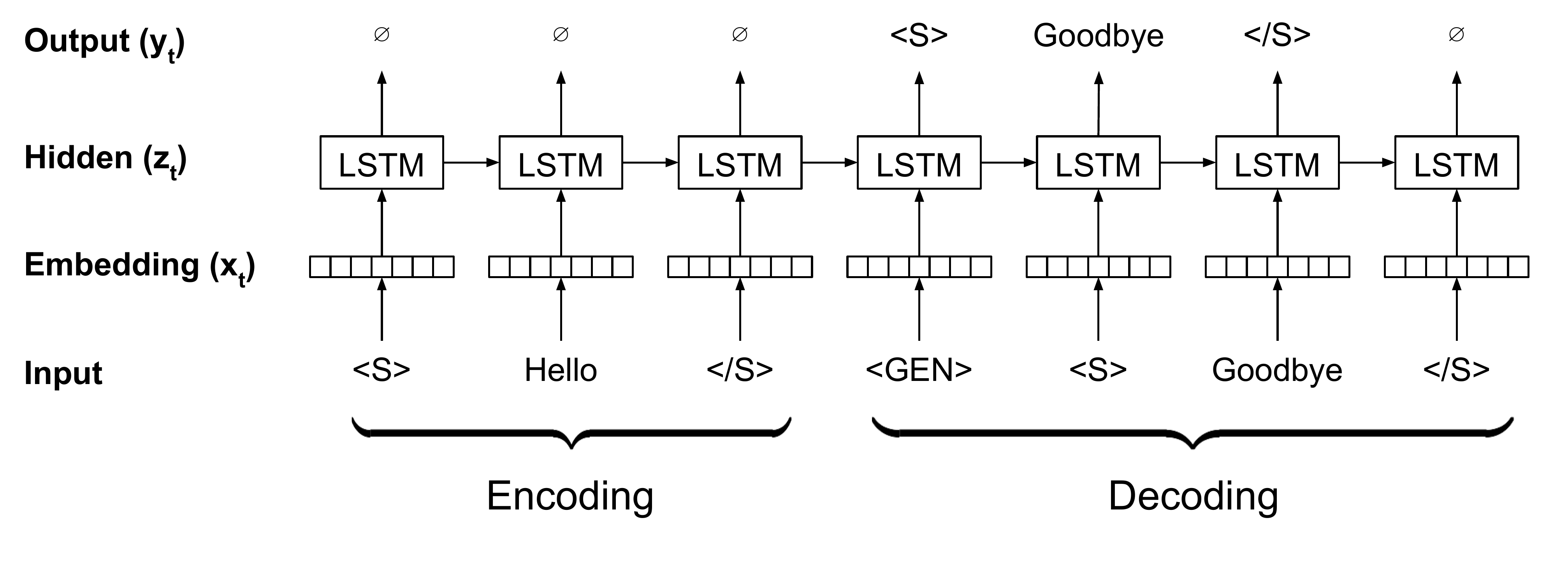}
\caption{Encoder-Decoder setup predicting the text ``Goodbye'' from ``Hello''
}
\label{fig:encoder}
\end{figure*}

Our general system architecture is as follows. At each timestep $t$, the input
token is represented as a learned 100-dimensional embedding vector (learned
jointly with the other parameters of the model), such that predictively similar
words should get similar embeddings. This embedding is fed as input to the LSTM
unit (that is, it will be the vector $x_t$ in Section \ref{sec:rnn}, the input
to the LSTM). The output of the LSTM unit (called $z_t$ in Section
\ref{sec:rnn}) is then fed to a softmax layer via a learned linear
transformation.

During the encoding phase the network is not trained to produce any output. During the decoding phase the output is a one-hot representation of the subsequent timestep's input
token (that is, with a $V$-word vocabulary, the output will be a $V$-dimensional
vector with one $1$ and $V-1$ zeros).
In this way, the network is trained to consume an entire input sequence and,
as a second step, iteratively output the subsequent timestep's input, which allows
the prediction of full output sequences.
This setup is pictured diagrammatically in Figure \ref{fig:encoder}, which gives
an example of input and output sequence for a token-level encoder-decoder model,
encoding the sentence ``Hello .'' and decoding the successor sentence 
``Goodbye .''
Note that we add beginning-of-sequence and end-of-sequence pseudo-tokens to 
sentences.
This formulation allows a system to be trained which can encode a sentence and 
then
infer a successor sentence by iteratively outputting next-input predictions 
until the
\texttt{</S>} end-of-sentence pseudo-token is predicted. We use different LSTMs
for encoding and decoding, as the dynamics of the two stages need not be 
identical.

We notate the different systems as follows. Let $s_1$ be the input sentence and
$s_2$ its successor sentence. Let $t_1$ denote the sequence of raw tokens in
$s_1$, and $t_2$ the tokens of $s_2$. Further, let $e_1$ and $e_2$ be the
sequence of structured events occurring in $s_1$ and $s_2$, respectively
(described in more detail in Section \ref{sec:expdetails}), and let $e_2[0]$
denote the first event of $e_2$. 
The different systems we
compare are named systematically as follows:
\begin{itemize}
  \item The system $t_1\shortrightarrow t_2$ is trained to encode a sentence's
  		tokens and decode its successor's tokens.
  \item The system $e_1\shortrightarrow e_2$ is trained to encode a sentence's
  		events and decode its successor's events.
  \item The system $e_1\shortrightarrow e_2 \shortrightarrow t_2$ is trained to
        encode a sentence's events, decode its successor's events, and then
        encode the latter and subsequently decode the successor's text.
\end{itemize}
We will not explicitly enumerate all systems, but other systems are
defined analogously, with the schema $X\shortrightarrow Y$ describing a system
which is trained to encode $X$ and subsequently decode $Y$, and $X
\shortrightarrow Y \shortrightarrow Z$ indicating a system which
is trained to encode $X$, decode $Y$, and subsequently encode $Y$ and decode 
$Z$. Note that
in a system $X\shortrightarrow Y \shortrightarrow Z$, only $X$ is provided as
input.

We also present results for systems of the form $X\attent Y$, which signifies
that the system is trained to decode $Y$ from $X$ with the addition of an 
attention mechanism. We use the attention mechanism of 
\newcite{vinyals:nips15}. In short, these models have additional parameters
which can learn soft alignments between positions of encoded inputs 
and positions in decoded outputs. Attention mechanisms have recently 
been shown to be quite empirically valuable in many complex sequence prediction
tasks. For more details on the model, see \newcite{vinyals:nips15}.

\begin{figure*}[t]
\centering
\includegraphics[width=0.95\textwidth]{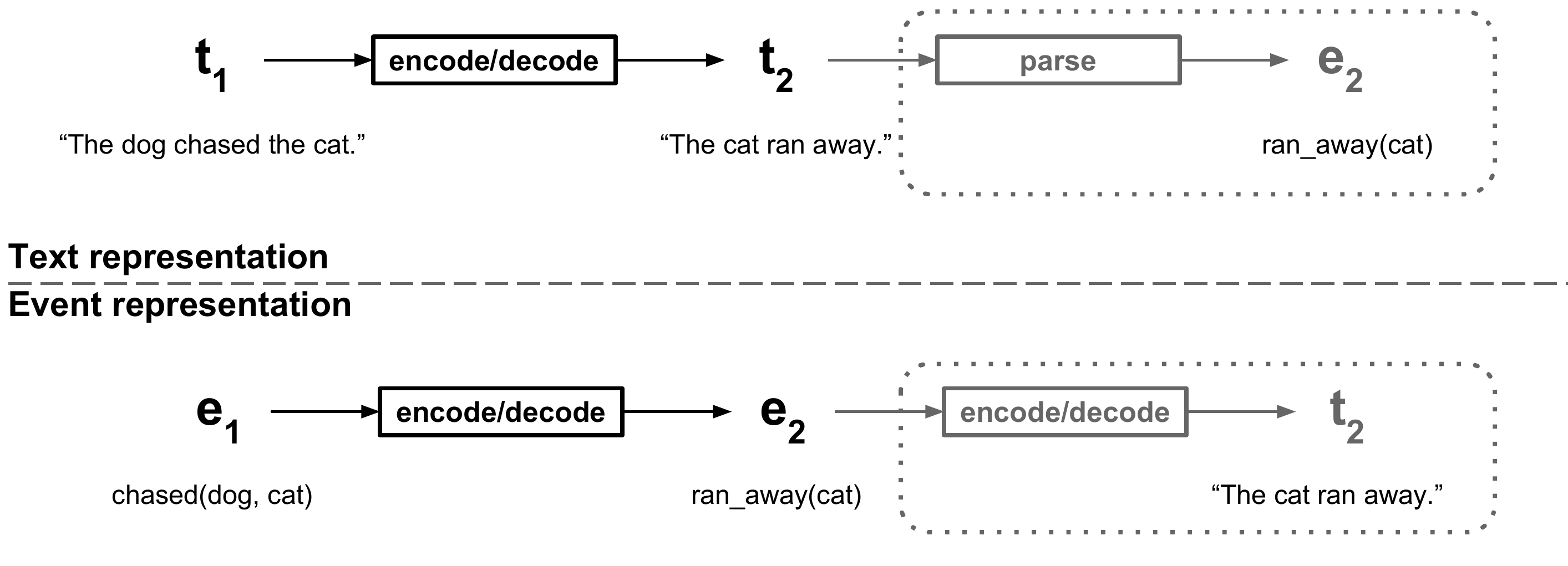}
\caption{Different system setups for modeling the two-sentence
sequence ``The dog chased the cat.'' followed by ``The cat ran away.'' The
gray components inside dotted boxes are only present in some systems.}
\label{fig:system}
\end{figure*}

Figure \ref{fig:system} gives a diagrammatic representation of the different
system setups. Text systems infer successor text and, optionally, parse that
text and extract events from it; event sequences infer successor events and,
optionally, expand inferred events into text.

Note that the system $t_1\shortrightarrow t_2$, in which both the
encoding and decoding steps operate on raw text, is essentially a
one-directional version of the skip-thought system of
\newcite{kiros:nips15}.\footnote{The system of \newcite{kiros:nips15}, in
addition to being trained to predict the next sentence, also contains a
backward-directional RNN trained to predict a sentence's predecessor; we condition only on previous text.
\newcite{kiros:nips15} also use Gated Recurrent Units instead of LSTM.}
Further, the system
$e_1\shortrightarrow e_2 \shortrightarrow t_2$,
which is trained to take a sentence's event sequence as input, predict its
successor's events, and then predict its successor's words, is comparable to the
event inference system of \newcite{pichotta:aaai16}. They use an LSTM
sequence model of events in sequence for event inference, and optionally
transform inferred events to text using another LSTM; we, on the other hand,
use an encoder/decoder setup to infer text directly.

\section{Evaluation}

\subsection{Experimental Details}
\label{sec:expdetails}

We train a number of LSTM encoder-decoder networks which vary in their input
and output. Models are trained on English Language Wikipedia, with 1\% of the
documents held out as a validation set. Our test set consists of 10,000 unseen
sentences (from articles in neither the training nor validation set). We train models with batch stochastic gradient descent with momentum, minimizing the cross-entropy error of
output predictions.
All models are implemented in TensorFlow \cite{abadi:tr15}.
We use a vocabulary of the 50,000 most frequent tokens, replacing all
other tokens with an out-of-vocabulary pseudo-token. Learned word embeddings are
100-dimensional, and the latent LSTM vector is 500-dimensional.
To extract events from text, we use the
Stanford Dependency Parser \cite{demarneffe:lrec06,socher:acl13}.
We use the Moses toolkit \cite{koehn:moses07} to calculate
\textsc{bleu}.\footnote{Via
the script \texttt{multi-bleu.pl}.}

We evaluate the task of predicting held-out text with
three metrics.
The first metric is \textbf{\textsc{bleu}}, which is standard
\textsc{bleu} (the geometric mean of modified 1-, 2-, 3-, and 4-gram precision
against a gold standard, multiplied by a brevity penalty which penalizes short
candidates). The second metric we present, \textbf{\textsc{bleu-bp},} is
\textsc{bleu} without the brevity penalty: in the task of predicting successor
sentences, depending on predictions' end use, on-topic brevity is not
necessarily undesirable. Evaluations are over top system inferences (that is, 
decoding is done by taking the argmax). Finally, we also present values for 
unigram precision (\textbf{1G P}), one of the components of \textsc{bleu}.

We also evaluate on the task of predicting held-out
verb-argument events, either directly or via inferred
text. We use two evaluation metrics for this task.
First, the \textbf{Accuracy} metric measures the percentage of a
system's most confident guesses that are totally correct. That is, for each
held-out event, a system makes its single most confident guess for that event,
and we calculate the total percentage of such guesses which are totally
correct. Some authors (e.g. \newcite{jans:eacl12}, \newcite{pichotta:aaai16})
present results on the ``Recall at $k$'' metric, judging gold-standard recall
against a list of top $k$ event inferences; this metric is equivalent to
``Recall at 1.'' This is quite a stringent metric, as an inference is only
counted correct if the verb and all arguments are correct. To relax this
requirement, we also present results on what we call the \textbf{Partial Credit}
metric, which is the percentage of held-out event components identical to the
respective components in a system's top inference.\footnote{This metric was used
in \newcite{pichotta:eacl14}, but there it was called \textit{Accuracy}. In the
present work, we use ``accuracy'' only to mean Recall at 1.} 

\subsection{Experimental Evaluation}

\begin{table}[htbp]
\begin{center}
\begin{tabular}{|l|l|l|l|}
  \hline
  \textbf{System} & \textbf{\textsc{bleu}}  & \textbf{\textsc{bleu-bp}} & \textbf{1G P} \\
  \hline\hline
  $t_1 \shortrightarrow t_1$
    & 1.88  & 1.88 & 22.6  \\
  \hline
  $e_1\shortrightarrow e_2 \shortrightarrow t_2$
    & 0.34  & 0.66 & 19.9  \\
  $e_1\attent e_2 \shortrightarrow t_2$
    & 0.30  & 0.39 & 15.8  \\
  \hline
  $t_1 \shortrightarrow t_2$ &
    \textbf{5.20}  & 7.84 & 30.9   \\
  $t_1 \attent t_2$ &
    4.68  & \textbf{8.09} & \textbf{32.2}   \\
  \hline
\end{tabular}
\caption{Successor text predictions evaluated with \textsc{bleu}.}
\label{tab:bleu}
\end{center}
\end{table}

Table \ref{tab:bleu} gives the results of evaluating predicted successor
sentence text against the gold standard using \textsc{bleu}.
The baseline system $t_1 \shortrightarrow t_1$ simply reproduces the input 
sentence as its own successor.\footnote{``$t_1 \shortrightarrow t_1$'' is minor abuse of notation,
as the system is not an encoder/decoder but a simple identity function.}
Below this are systems which make predictions from event information, with
systems which make predictions from raw text underneath. Transformations
written $X \attent Y$ are, recall, encoder-decoder LSTMs with 
attention.

Note, first, that the text-level models outperform other models on 
\textsc{bleu}. In particular, the two-step model $e_1\shortrightarrow e_2 
\shortrightarrow t_2$ (and comparable model with attention) which first
predicts successor events and then, as a separate step, expands these events
into text, performs quite poorly. This is perhaps due to the fact that the
translation from text to events is lossy, so reconstructing raw sentence tokens
is not straightforward.

The \textsc{bleu-bp} scores, which are \textsc{bleu} without the brevity 
penalty, are noticeably higher in the text-level models than the raw 
\textsc{bleu} scores. This is in part because these models seem to 
produce shorter sentences, as illustrated below in section 
\ref{sec:qualitative}.

The attention mechanism does not obviously benefit either text or event level
prediction encoder-decoder models. This could be because there is not 
an obvious alignment structure between contiguous spans of raw text (or events)
in natural documents.

These results provide evidence that, if the Narrative Cloze task is defined to
evaluate prediction of held-out text from a document, then sentence-level RNN
language models provide superior performance to RNN models operating
at the event level. In other words, linguistic pre-processing does not
obviously benefit encoder-decoder models trained to predict succeeding text.

\begin{table}[ht]
\begin{center}
\begin{tabular}{|l|l|l|}
\hline
  \textbf{System} & \textbf{Accuracy}  & \textbf{Partial Credit}  \\
  \hline\hline
  Most common & 0.2 & 26.5 \\
  \hline
  $e_1\shortrightarrow e_2[0] $ &
    \textbf{2.3}  & 26.7  \\
  $e_1\attent e_2[0] $ &
    2.2  & 25.6  \\
    \hline
  $t_1 \shortrightarrow t_2 \shortrightarrow e_2[0]$ &
    2.0  & \textbf{30.3}   \\
  $t_1 \attent t_2 \shortrightarrow e_2[0]$ &
    2.0  & 27.7   \\
  \hline
\end{tabular}
\caption{Next event prediction accuracy (numbers are percentages: maximum value
is 100).}
\label{tab:cloze}
\end{center}
\end{table}
Table \ref{tab:cloze} gives results on the task of predicting the next verb
with its nominal arguments;
that is, whereas Table \ref{tab:bleu} gave results on a text analog to
the Narrative Cloze evaluation (\textsc{bleu}), Table \ref{tab:cloze} gives 
results on the verb-with-arguments prediction version. 
In the $t_1 \shortrightarrow t_2 \shortrightarrow e_2[0]$ system (and the 
comparable system with attention), events are 
extracted from automatically generated text by parsing output text and applying 
the same event extractor to this parse used to extract events from raw 
text.\footnote{This is also a minor abuse of notation, as the second 
transformation uses a statistical parser rather than an encoder/decoder.}
The row labeled \textbf{Most common} in Table \ref{tab:cloze} gives performance for
the baseline system which always guesses the most common event in the training
set.

The LSTM models trained to directly predict events are roughly comparable 
to systems which operate on raw text, performing slightly worse on accuracy 
and slightly better when taking partial credit into account. As with the
previous comparisons with \textsc{bleu}, the attention mechanism does not
provide an obvious improvement when decoding inferences, perhaps, again, because
the event inference problem lacks a clear alignment structure.

These systems infer their most probable guesses of $e_2[0]$, the first event
in the succeeding sentence. In order for a system prediction to be counted as 
correct, it must have the correct strings for grammatical head words of all 
components of the correct event. Note also that we judge only against a 
system's single most confident prediction (as opposed to some prior work
\cite{jans:eacl12,pichotta:eacl14} which takes the top $k$ predictions---the
numbers presented here are therefore noticeably lower). We
do this mainly for computational reasons: namely, a beam search over a full
sentence's text would be quite computationally expensive.

\subsection{Adding Additional Context}
\label{sec:context}

The results given above are for systems which encode information about one 
sentence and decode information about its successor. This is within the spirit
of the skip-gram system of \newcite{kiros:nips15}, but we may wish to 
condition on more of the document. 
To investigate this, we perform an experiment varying the number of previous 
sentences input during the encoding step of $t_1 \shortrightarrow t_2$ 
text-level models without attention. We train three different models, which 
take either one, three, or five sentences as input, respectively, and are 
trained to output the successor sentence.

\begin{table}[htbp]
\begin{center}
\begin{tabular}{|l|l|l|l|}
  \hline
  \textbf{Num Prev Sents} & \textbf{\textsc{bleu}}  & \textbf{\textsc{bleu-bp}} & \textbf{1G P} \\
  \hline\hline
  1 & 5.80  & 8.59 & 29.4  \\
  3 & 5.82  & \textbf{9.35} & \textbf{31.2}  \\
  5 & \textbf{6.83}  & 6.83 & 21.4  \\
  \hline
\end{tabular}
\caption{Varying the amount of context in text-level models. ``Num Prev 
Sents'' is the number of previous sentences supplied during encoding.}
\label{tab:nback}
\end{center}
\end{table}

Table \ref{tab:nback} gives the results of running these models on 10,000 
sentences from the validation set. As can be seen, in the training
setup we investigate, more additional context sentences have a mixed effect,
depending on the metric. This is perhaps due in 
part to the fact that we kept hyperparameters fixed between experiments, and
a different hyperparameter regime would benefit predictions from longer input 
sequences. More investigation could prove fruitful.


\subsection{Qualitative Analysis}
\label{sec:qualitative}

\begin{figure*}[t]
\centering
\begin{small}
\begin{tabular}{l p{120mm}}
\hline
\textbf{Input}: & As of October 1 , 2008 , $\langle$OOV$\rangle$ changed its company name
                to Panasonic Corporation. \\
\textbf{Gold}: & $\langle$OOV$\rangle$ products that were branded ``National'' in Japan
            are currently marketed under the ``Panasonic'' brand. \\
\textbf{Predicted}: & The company's name is now $\langle$OOV$\rangle$. \\
\hline
\textbf{Input}: & White died two days after Curly Bill shot him. \\
\textbf{Gold}: & Before dying, White testified that he thought the pistol had accidentally discharged and that he did not believe that Curly Bill shot him on purpose. \\
\textbf{Predicted}: & He was buried at $\langle$OOV$\rangle$ Cemetery. \\
\hline
\textbf{Input}: & The foundation stone was laid in 1867. \\
\textbf{Gold}: & The members of the predominantly Irish working class parish
            managed to save \pounds 700 towards construction, a large sum at the 
            time. \\
\textbf{Predicted}: & The $\langle$OOV$\rangle$ was founded in the early 20th century. \\
\hline
\textbf{Input}: & Soldiers arrive to tell him that $\langle$OOV$\rangle$ has been seen in
                camp and they call for his capture and death. \\
\textbf{Gold}: & $\langle$OOV$\rangle$ agrees . \\
\textbf{Predicted}: & $\langle$OOV$\rangle$ is killed by the $\langle$OOV$\rangle$. \\
\hline
\end{tabular}
\end{small}

\caption{Sample next-sentence text predictions. $\langle$OOV$\rangle$ is the
out-of-vocabulary pseudo-token, which frequently replaces proper names.}
\label{fig:examples}
\end{figure*}

Figure \ref{fig:examples} gives some example automatic next-sentence text 
predictions, along with the input sentence and the gold-standard next sentence.
Note that gold-standard successor sentences frequently introduce new details 
not obviously inferrable from previous text. Top system predictions,
on the other hand, are frequently fairly short. This is likely due part to
the fact that the cross-entropy loss does not directly penalize short 
sentences and part to the fact that many details in gold-standard successor
text are inherently difficult to predict.

\subsection{Discussion}

The general low magnitude of the
\textsc{bleu} scores presented in Table \ref{tab:bleu}, especially in comparison
to the scores typically reported in Machine Translation results, indicates the
difficulty of the task. In open-domain text, a sentence is typically not
straightforwardly predictable from preceding text; if it were, it would likely
not be stated.

On the task of verb-argument prediction in Table \ref{tab:cloze}, the difference
between $t_1 \shortrightarrow t_2$ and $e_1 \shortrightarrow e_2[0]$ is
fairly marginal. This raises the general question of how
much explicit syntactic analysis is required for the task of
event inference, particularly in the encoder/decoder setup. These results provide evidence that a sentence-level RNN
language model which operates on raw tokens can predict what comes next in a
document as well or nearly as well as an event-mediated script model.

\section{Future Work}

There are a number of further extensions to this work. First, in this work (and,
more generally, Neural Machine Translation research), though generated text is
evaluated using \textsc{bleu}, systems are optimized for per-token
cross-entropy error, which is a different objective (\newcite{luong:iclr16}
give an example of a system which improves cross-entropy error but
reduces \textsc{bleu} score in the Neural Machine Translation context).
Finding differentiable objective functions that more directly target more complex
evaluation metrics like \textsc{bleu} is an interesting future
research direction.

Relatedly, though we argue that \textsc{bleu} is a natural token-sequence-level
analog to the verb-argument formulation of the Narrative Cloze task, it is not
obviously the best metric for evaluating inferences of text, and comparing these
automated metrics with human judgments is an important direction of future work.
\newcite{pichotta:aaai16} present results on crowdsourced human evaluation of
script inferences that could be repeated for our RNN models.

Though we focus here on forward-direction models predicting successor
sentences, bidirectional encoder-decoder models, which predict sentences from
both previous and subsequent text, are another interesting future research
direction.

\section{Related Work}

The use of scripts in AI dates back to the 1970s
\cite{minsky:technote74,schank:book77}; in this conception, scripts were
composed of complex events with no probabilistic semantics, which were difficult
to learn automatically. In recent years, a growing body of research has
investigated learning probabilistic co-occurrence models with simpler
events. \newcite{chambers:acl08} propose a model of co-occurrence of
(verb, dependency) pairs, which can be used to infer such pairs from documents;
\newcite{jans:eacl12} give a superior model in the same general framework.
\newcite{chambers:acl09} give a method of generalizing from single sequences
of pair events to collections of such sequences.
\newcite{rudinger:emnlp15} apply a discriminative language model to the
(verb, dependency) sequence modeling task, raising the question of to what
extent event inference can be performed with standard language models
applied to event sequences. \newcite{pichotta:eacl14} describe a method of
learning a co-occurrence based model of verbs with multiple
coreference-based entity arguments.

There is a body of related work focused on learning models of co-occurring
events to automatically induce templates of complex
events comprising multiple verbs and arguments, aimed ultimately at maximizing
coherency of templates
\cite{chambers:emnlp13,cheung:naacl13,balasubramanian:emnlp13}.
\newcite{ferraro:aaai16} give a model integrating various levels of event
information of increasing abstraction, evaluating both on coherence of induced
templates and log-likelihood of predictions of held-out events.
\newcite{mcintyre:acl10} describe a system that learns a model of co-occurring
events and uses this model to automatically generate stories via a Genetic
Algorithm.

There have been a number of recent published neural models for various event-
and discourse-related tasks. \newcite{pichotta:aaai16} show that an LSTM
event sequence model outperforms previous co-occurrence methods for predicting
verbs with arguments. \newcite{granroth-wilding:aaai16} describe a feedforward
neural network which composes verbs and arguments into low-dimensional vectors,
evaluating on a multiple-choice version of the Narrative Cloze task.
\newcite{modi:conll14} describe a feedforward network which is trained to
predict event orderings.
\newcite{kiros:nips15} give a method of embedding sentences in low-dimensional
space such that embeddings are predictive of neighboring sentences.
\newcite{li:emnlp14b} and \newcite{ji:tacl15}, use RNNs for discourse parsing;
\newcite{liu:aaai16} use a Convolutional Neural Network for implicit discourse
relation classification.

\section{Conclusion}

We have given what we believe to be the first systematic evaluation of
sentence-level RNN language models on the task of predicting held-out document 
text.
We have found that models operating on raw text perform
roughly comparably to identical models operating on  
predicate-argument event structures when predicting the latter, and that text
models provide superior predictions of raw text. This provides
evidence that, for the task of held-out event prediction, encoder/decoder
models mediated by automatically extracted events may not be learning 
appreciably more structure than systems trained on raw tokens alone.

\section*{Acknowledgments}
Thanks to Stephen Roller, Amelia Harrison, and the UT NLP group for their help 
and feedback.
Thanks also to the anonymous reviewers for their very helpful suggestions.
This research was supported in part by the DARPA DEFT program under AFRL
grant FA8750-13-2-0026.

\bibliography{local}
\bibliographystyle{acl2016}

 \appendix
 \section{Supplemental Material}
 \label{sec:supplemental}

 Our Wikipedia dump from which the training, development, and test sets are
 constructed is from Jan 2, 2014. We parse text using version 3.3.1 of the
 Stanford CoreNLP system.
 We use a vocab consisting of the 50,000 most common tokens, replacing all others
 with an Out-of-vocabulary pseudotoken. We train using batch stochastic gradient
 descent with momentum with a batch size of 10 sequences, using an initial
 learning rate of 0.1, damping the learning rate by 0.99 any time the previous
 hundred updates' average test error is greater than any of the average losses
 in the previous ten groups of hundred updates.
 Our momentum parameter is 0.95. Our embedding vectors are
 100-dimensional, and our LSTM hidden state is 500-dimensional. We train all 
 models for 300k batch updates (with the exception of the models compared in 
 \S\ref{sec:context}, all of which we train for 150k batch updates, as training 
 is appreciably slower with longer input sequences). Training takes 
 approximately 36 hours on an NVIDIA Titan Black GPU.

%
%
%


\end{document}